\documentclass[runningheads]{llncs}

\usepackage{comment}
\usepackage{graphicx}
\usepackage{amsmath}
\usepackage{amssymb}
\usepackage{amsfonts}
\usepackage{microtype}
\usepackage{makecell}
\usepackage{multirow}
\usepackage{array}
\usepackage{hyperref}
\hypersetup{
    colorlinks=true
}
\renewcommand{\vec}[1]{\boldsymbol{#1}}
\newcommand{\mat}[1]{\mathbf{#1}}
\newcommand{\set}[1]{\mathcal{#1}}

\newcommand{\pose}[0]{\vec{\theta}}
\newcommand{\shape}[0]{\vec{\beta}}

\newcommand{\enc}[0]{f_\mathrm{enc}}
\newcommand{\dec}[0]{g_\mathrm{dec}}

\makeatletter
\newcommand{\putindeepbox}[2][0.7\baselineskip]{{%
    \setbox0=\hbox{#2}%
    \setbox0=\vbox{\noindent\hsize=\wd0\unhbox0}
    \@tempdima=\dp0
    \advance\@tempdima by \ht0
    \advance\@tempdima by -#1\relax
    \dp0=\@tempdima
    \ht0=#1\relax
    \box0
}}
\makeatother

\makeatletter
\newcommand{\thickhline}{%
    \noalign {\ifnum 0=`}\fi \hrule height 1pt
    \futurelet \reserved@a \@xhline
}
\makeatother

\begin{document}
% \linenumbers
\pagestyle{headings}
\mainmatter
\def\ECCVSubNumber{4065}  % Insert your submission number here

\title{Unsupervised Shape and Pose Disentanglement for 3D Meshes}

%\begin{comment}
\titlerunning{}
\author{Keyang Zhou \and
Bharat Lal Bhatnagar \and
Gerard Pons-Moll}
\authorrunning{}
\institute{Max Planck Institute for Informatics, Saarland Informatics Campus, Germany
\email{\{kzhou,bbhatnag,gpons\}@mpi-inf.mpg.de}}

%\end{comment}
%******************
\maketitle

\begin{abstract}
Parametric models of humans, faces, hands and animals have been widely used for a range of tasks such as image-based reconstruction, shape correspondence estimation, and animation. Their key strength is the ability to factor surface variations into shape and pose dependent components.
Learning such models requires lots of expert knowledge and hand-defined object-specific constraints, making the learning approach unscalable to novel objects. In this paper, we present a simple yet effective approach to learn disentangled shape and pose representations in an unsupervised setting. We use a combination of self-consistency and cross-consistency constraints to learn pose and shape space from registered meshes. We additionally incorporate as-rigid-as-possible deformation(ARAP) into the training loop to avoid degenerate solutions. We demonstrate the usefulness of learned representations through a number of tasks including pose transfer and shape retrieval. The experiments on datasets of 3D humans, faces, hands and animals demonstrate the generality of our approach. Code is made available at \url{https://virtualhumans.mpi-inf.mpg.de/unsup_shape_pose/}.

\keywords{3D Deep Learning, Disentanglement, Body Shape, Mesh Auto-encoder, Representation Learning}
\end{abstract}

\section{Introduction}
\begin{figure}[ht]
	\centering
	\includegraphics[width=\textwidth]{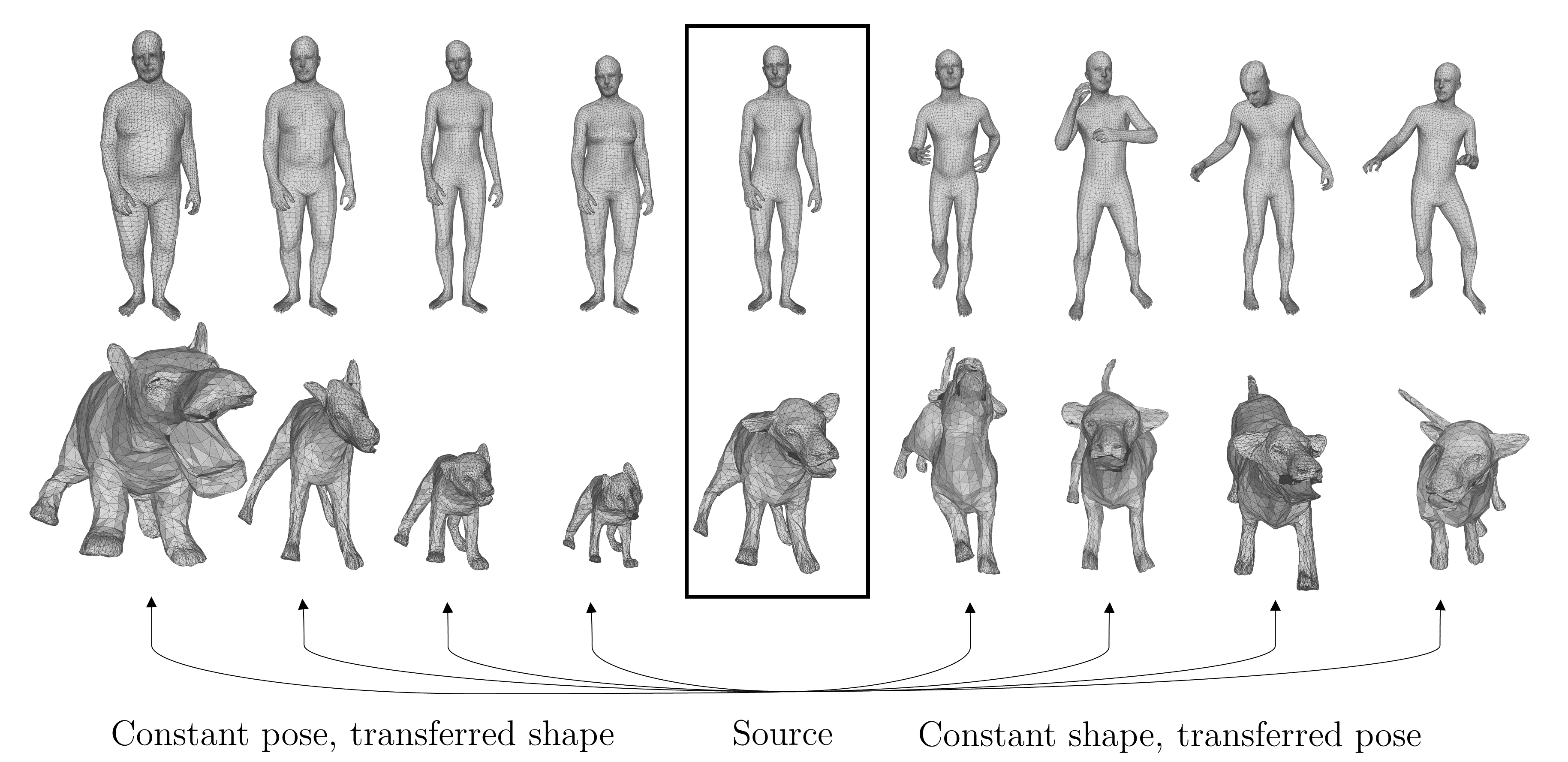}

	\caption{Our model learns a disentangled representation of shape and pose for mesh. In the middle are two source subjects taken from AMASS and SMAL datasets respectively. On the left are meshes with the same pose but varying shapes which we construct by transferring shape codes extracted from other meshes using our method. On the right are meshes with the same subject identity but varying poses which we construct by transferring pose codes.}
	\label{fig:teaser}
\end{figure}

Parameterizing 3D mesh deformation with different factors, such as pose and shape, is crucial in computer graphics for efficient 3D shape manipulation, and for computer vision, to extract structure and understand human and animal motion in videos.

Although parametric models of meshes such as SCAPE~\cite{anguelov2005scape}, SMPL~\cite{loper2015smpl}, Dyna~\cite{ponsmollSIGGRAPH15Dyna}, Adam~\cite{joo2018total} for bodies, MANO~\cite{romero2017embodied} for hands, SMAL~\cite{zuffi20173d} for animals, basel face model~\cite{bfm09}, FLAME~\cite{FLAME:2017} and their combinations~\cite{ploumpis2019combining} for faces, have been extremely useful for many applications. Learning them is a \emph{difficult task} that requires expert knowledge and manual intervention. SMPL for example, is learned from a set of meshes in correspondence, and requires defining a skeleton hierarchy, manually initializing blendweights to bind each vertex to body parts, carefully \emph{unposing} meshes, and a training procedure that requires several stages. 

In this paper, we address the problem of unsupervised disentanglement of pose and shape for 3D meshes. 
Like other models such as SMPL, our method requires a dataset of meshes registered to a template for training. But unlike other methods, we learn to factor pose and shape based on the data alone without making assumptions on the number of parts, the skeleton or the kinematic chain. Our model only requires that the same shape can be seen in different poses, which is available for datasets collected from scanners or motion capture devices. We call our model unsupervised because we do not make use of meshes annotated with pose or shape codes, and we make no assumptions on the underlying parts or skeleton. This flexibility makes our model applicable to a wide variety of objects, such as humans, hands, animals and faces.  

\begin{figure}[t]
	\centering
	\includegraphics[width=\textwidth]{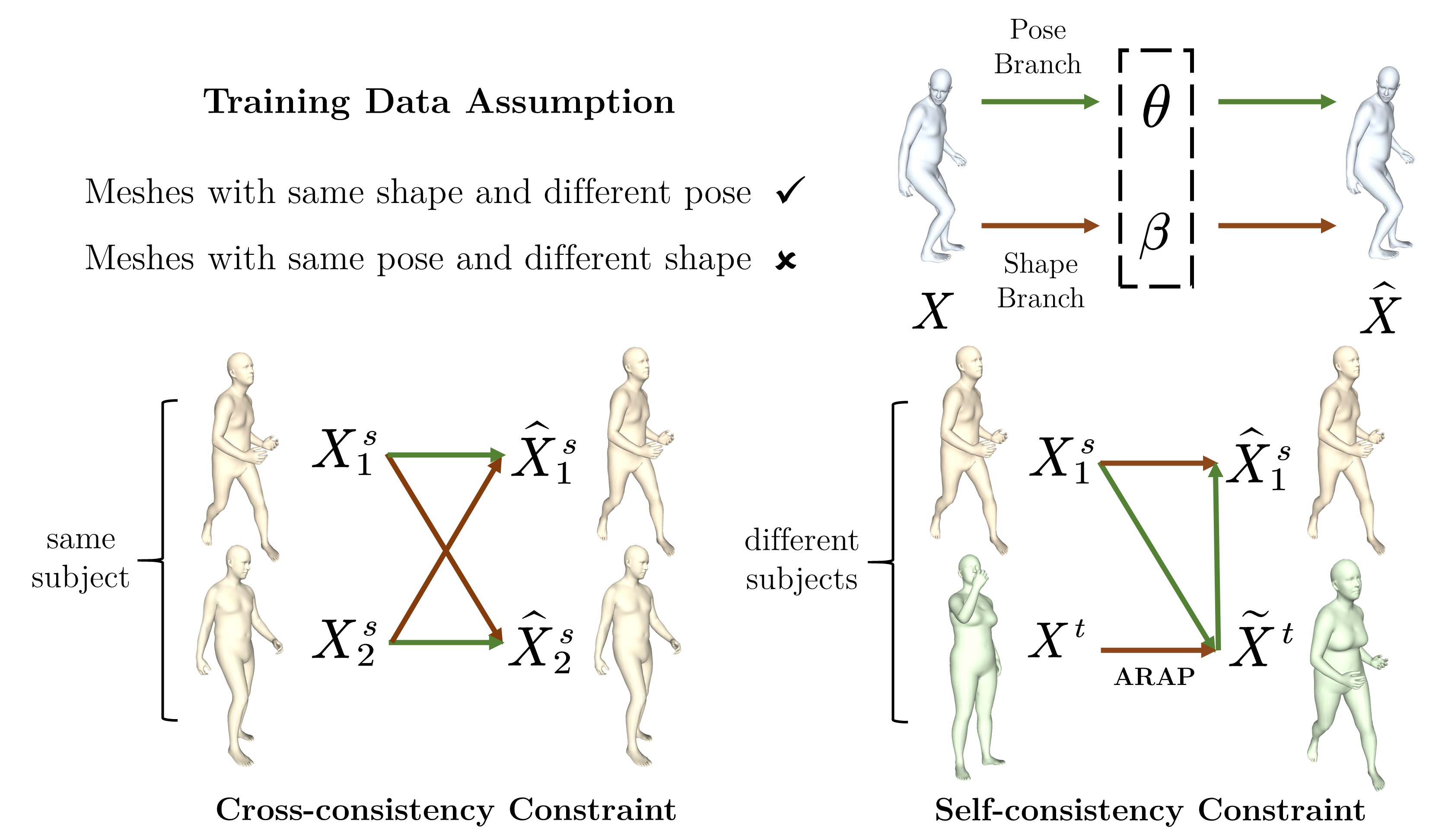}
	\caption{A schematic overview of shape and pose disentangling mesh auto-encoder. The input mesh $\mat{X}$ is separately processed by a shape branch and a pose branch to get shape code $\vec{\beta}$ and pose code $\vec{\theta}$. The two latent codes are subsequently concatenated and decoded to the reconstructed mesh $\mat{\hat{X}}$(top). The shape codes of two deformations of the same subject are swapped to reconstruct each other(bottom left). The pose code of one subject is used to reconstruct itself after a cycle of decoding-encoding(bottom right).}
	\label{fig:overview}
\end{figure}

Unsupervised disentanglement from meshes is a challenging task. 
Most datasets~\cite{mahmood2019amass,FLAME:2017,romero2017embodied,loper2015smpl} contain the same shape in different poses, \emph{e.g.}, they capture a human or an animal moving. However, real world datasets \emph{do not contain two different shapes in the same pose} -- two different humans, or animals are highly unlikely to be captured performing the exact same pose or motion. 
This makes disentangling pose and shape from data difficult.

We achieve disentanglement with an auto-encoding neural network based on two key observations.
First, we should be able to auto-encode a mesh in two codes (pose and shape), which we achieve with two separate encoder branches, see Fig.~\ref{fig:overview}(top).
 Second, given two meshes $\mat{X}_{1}^{s}$ and $\mat{X}_{2}^{s}$ of the same subject $s$ in two different poses, we should be able to swap their shape codes and reconstruct exactly the two input meshes. This is imposed with a \emph{cross-consistency loss}, see Fig.~\ref{fig:overview}(lower left). These two constraints however, are not sufficient and lead to degenerate solutions, with shape information flowing into pose code. 

If we had access to two different shapes in the exact same pose, we could impose an analogous cross-consistency loss on the pose. But as mentioned, such data is not available. Our idea is to \emph{generate} such pairs of \emph{different shapes} with the \emph{exact same pose} on the fly during training with our disentangling network. 

Given two meshes with different shapes and poses $\mat{X}_1^s$ and $\mat{X}^t$, we generate a proxy mesh $\tilde{\mat{X}}^{t}$ with the pose of mesh $\mat{X}^{s}_1$ and the shape of mesh $\mat{X}^t$ within the training loop. If disentanglement is effective, we should recover the original pose code from the proxy mesh, and mix it with the shape code of mesh $\mat{X}_1^s$, to decode it into mesh $\mat{X}_{1}^{s}$.
We ask the network to satisfy this constraint with a \emph{self-consistency loss}.
For the self-consistency constraint to work well, the proxy mesh must not contain any shape characteristic of mesh $\mat{X}_1^s$, which occurrs if the pose code carries shape information. 
To resolve this, we replace the initially decoded proxy mesh $\mat{\tilde{X}}^t$ with an As-Rigid-As-Possible~\cite{sorkine2007rigid} approximate. 
Self-consistency is best understood with the illustration in Fig.~\ref{fig:overview}(lower right).
 
Our experiments show that these two simple---but not immediately obvious---losses allow to discover independent pose and shape factors from 3D meshes directly. 
To demonstrate the wide applicability of our method, we use it to disentangle pose and shape in four different publicly available datasets of full body humans~\cite{mahmood2019amass}, hands~\cite{romero2017embodied}, faces~\cite{FLAME:2017} and animals~\cite{zuffi20173d}. 
We show several downstream applications, such as pose transfer, pose-aware shape retrieval, and pose and shape interpolation.  
We will make our code and model publicly available so that researchers can learn their own models from data.

\section{Related Work}
\subsubsection{Disentangled representations for 2D images.} The motivation behind feature disentanglement is that images can be synthesized from individual factors of variation. A pioneering work for disentanglement learning is InfoGAN~\cite{chen2016infogan}, which maximizes the variational lower bound for the mutual information between latent code and generator distribution. Beta-VAE~\cite{higgins2017beta} and its follow-up work~\cite{chen2018isolating} penalized a KL divergence term to reduce variable correlations. Similarly, Kim et al.~\cite{kim2018disentangling} encouraged fatorial marginal distribution of latent variables.

Another line of work incorporates Spatial Transformer Network~\cite{jaderberg2015spatial} to explicitly model object deformations~\cite{shu2018deforming,detlefsen2019explicit,lorenz2019unsupervised}. Iosanos et al.~\cite{sahasrabudhe2019lifting} recovered a 3D deformable template from a set of images and transformed it to fit image coordinates. Recently, adversarial training is exploited to enforce feature disentanglement~\cite{liu2018unified,esser2019unsupervised,szabo2017challenges,mathieu2016disentangling,denton2017unsupervised}. Our work has similarities with~\cite{hu2018disentangling,zhang2019multi}, where latent features are mixed and then separated. But unlike them, our method does not depend on auxiliary classifiers or adversarial loss, which are notoriously hard to train and tune. The idea of swapping codes (cross-consistency) to factor out appearance or identity as been also used in ~\cite{rhodin2018unsupervised}, but we additionally introduce the self-consistency loss which is critical for disentanglement. Furthermore, all these works focus on 2D images while we focus on disentanglement for 3D meshes. 

\subsubsection{Deep learning for 3D reconstructions.} With the advances in geometric deep learning, a number of models have been proposed to analyse and reconstruct 3D shapes. Particularly related to us are mesh auto-encoders. Tan et al.~\cite{tan2018variational} designed a mesh variational auto-encoder using fully-connected layers. Instead of operating directly on mesh vertices, the model deals with a rotation-invariant mesh representation~\cite{gao2016efficient}. Ranjan et al.~\cite{ranjan2018generating} generalized downsampling and upsampling layers to meshes by collapsing unimportant edges based on quadric error measure. DEMEA~\cite{tretschk2019demea} performs mesh deformation in a low-dimensional embedded deformation layer which helps reduce reconstruction artifacts. These models do not separate shapes from poses when embedding meshes into the latent space. Jiang et al.~\cite{jiang2019disentangled} decomposed 3D facial meshes into identity code and expression code. Their approach needs supervision on expression labels to work. Similarly, Jiang et al.~\cite{9072593} trained a disentangled human body model in a hierarchical manner with a predefined anatomical segmentation. 
Deng et al.~\cite{deng2019neural} conditions human shape occupancy on pose, but requires pose labels for training.
Levinson et al.~\cite{levinson2019latent} trained on pairs of shapes with the exact same poses, which is unrealistic for non-synthetic datasets. LIMP~\cite{cosmo2020limp} explicitly enforced that change in pose should preserve pairwise geodesic distances. Although it works well for small datasets, the intensive computations make it unsuitable for larger datasets. Geometrically Disentangled VAE(GDVAE)~\cite{aumentado2019geometric} is capable of learning shape and pose from pointclouds in a completely unsupervised manner. GDVAE utilizes the fact that isometric deformations preserve spectrum of the Laplace-Beltrami Operator(LBO) to disentangle shape. While we require meshes in correspondence and GDVAE does not, we obtain significantly better disentanglement and reconstruction quality. Furthermore, in practice GDVAE uses meshes in correspondence to compute the LBO spectrum of each mesh. While the spectrum should be invariant to connectivity, in practice it is known to be very sensitive to noise and different discretizations. 
Instead of relying on LBO spectrum, we assume the subject identity is known which requires no extra labelling, and impose shape and pose consistency by swapping and mixing codes during training. 

\subsubsection{3D deformation transfer.} Traditional deformation transfer methods solve an optimization problem for each pair of source and target meshes. The seminal work of Sumner et al.~\cite{sumner2004deformation} transfers deformation via per-triangle affine transformations assuming correspondence. While general, this approach produces artifacts when transferring between significantly different shapes.
Ben-Chen et al.~\cite{ben2009spatial} formulated deformation transfer as a space deformation problem. Recently, Lin et al.~\cite{gao2018automatic} achieved automatic deformation transfer between two different domains of meshes without correspondence. They build an auto-encoder for each of the source and target domain. Deformation transfer is performed at latent space by a cycle-consistent adversarial network~\cite{zhu2017unpaired}. For every new pair of shapes, a new model needs to be trained, whereas we train on multiple shapes simultaneously, and our training procedure is much simpler.
These approaches focus on transferring pose deformations between pairs of meshes, whereas our ability to transfer deformation is just a natural consequence of the learned disentangled representation.

\section{Method}
Given a set of meshes with the same topology, our goal is to learn a latent representation with disentangled shape and pose components. In our context, we refer to shape as the intrinsic geometric properties of a surface (height, limb lengths, body shape etc.), which remain invariant under approximately isometric deformations . 
We refer to the other properties that vary with motion as pose.

Our model is built on three mild assumptions. i) All the meshes should be registered and have the same connectivity. ii) There are enough shape and pose variations in the training set to cover the latent space. iii) The same shape can be seen in different poses, which naturally occurs when capturing a body, face, hand or animal in motion. 
Note that models like SMPL~\cite{loper2015smpl} are built on the same assumptions, but unlike those models we do not hand-define the number of parts, skeleton nor the surface-to-part associations.

\subsection{Overview}
Our model follows the classical auto-encoder architecture. The encoder function $\enc$ embeds input mesh $\mat{X}$ into latent shape space and latent pose space: $\enc(\mat{X}) = \left(f_{\shape}(\mat{X}), f_{\pose}(\mat{X})\right)= (\vec{\beta}, \vec{\theta})$, where $\vec{\beta}$ denotes shape code, and $\vec{\theta}$ denotes pose code. 
The encoder consists of two branches for shape $f_{\vec{\beta}}(\mat{X})= \vec{\beta}$ and for pose $f_{\vec{\theta}}(\mat{X})=\vec{\theta}$ respectively, which are independent and do not share weights. The decoder function $g_\mathrm{dec}$ takes shape and pose codes as inputs, and transforms them back to the corresponding mesh: $\dec(\vec{\beta}, \vec{\theta})=\tilde{\mat{X}}$.

The challenge is to disentangle pose and shape in an unsupervised manner, without supervision on $\vec{\theta}$ or $\vec{\beta}$ coming from an existing parametric model.
We achieve this with a cross-consistency and a self-consistency loss during training. 
An overview of our approach is given in Fig.~\ref{fig:overview}. 

\subsection{Cross-consistency}
Given two meshes, $\mat{X}_{1}^{s}$ and $\mat{X}_{2}^{s}$ (superscript indicates subject identity and subscript labels individual meshes of a given subject), of subject $s$ in different poses we should be able to swap their shape codes and recover exactly the same meshes. 
% Suppose $\mat{X}_{1}^{s}$ and $\mat{X}_{2}^{s}$ are two meshes of subject $s$ in different poses
% (superscripts indicate subject identities and subscripts label individual meshes of a given subject). Ideally we should be able to swap their shape codes and recover exactly the same meshes. 

We randomly sample a mesh pair $(\mat{X}_1^s, \mat{X}_2^s)$ of the same subject from the training set and decompose it into $(\vec{\beta}_{1}^{s}, \vec{\theta}_{1}^{s})$ and $(\vec{\beta}_{2}^{s}, \vec{\theta}_{2}^{s})$ respectively. The cross-consistency implies that the original meshes should be recovered by swapping shape codes $\vec{\beta}_{1}^{s}$ $\vec{\beta}_{2}^{s}$:
\begin{align}
    \dec(\vec{\beta}_{2}^{s}, \vec{\theta}_{1}^{s})&=\mat{X}_{1}^{s} \\
    \dec(\vec{\beta}_{1}^{s}, \vec{\theta}_{2}^{s})&=\mat{X}_{2}^{s}
\end{align}
Since the cross-consistency constraint holds in both directions, optimizing one loss term suffices. The loss is defined as
\begin{equation}
\mathcal{L}_C = {\left\lVert {\dec\left(f_{\vec{\beta}}(\mat{X}_2^s), f_{\vec{\theta}}(\mathcal{T}(\mat{X}_1^s))\right) - \mat{X}_1^s} \right\rVert}_{1},
\end{equation}
where $\mathcal{T}$ is a family of pose invariant mesh transformations such as random scaling and uniform noise corruption, which serves as data augmentation to improve generalization and robustness of the pose branch.  
The cross-consistency is useful to make the model aware of the distinction between shape and pose, but as we discussed in the introduction, it alone does not guarantee disentangled representations. This motivates our self-consistency loss, which we explain next.

\subsection{Self-consistency}
Having pairs of meshes with different shapes and the exact same pose would simplify the task, but such data is never available in real world datasets. 
The key idea of self-consistency is to generate such mesh pairs consisting of two different shapes in the same pose on the fly during the training process. 

We sample a triplet $(\mat{X}_{1}^{s}, \mat{X}_{2}^{s}, \mat{X}^{t})$, where mesh $\mat{X}^t$ shares neither shape nor pose with $(\mat{X}_1^s, \mat{X}_2^s)$. 
We combine the shape from $\mat{X}^t$ and pose from $\mat{X}_1^s$ to generate an intermediate mesh $\mat{\tilde{X}}^t = \dec(\vec{\beta}^t, \vec{\theta}_1^s)$. 

Since $\mat{\tilde{X}}^t$ should have the same pose $\vec{\tilde{\theta}}^{t} = f_{\vec{\theta}}(\mat{\tilde{X}}^t)$ as $\mat{X}_1^s$, and $\mat{X}_2^s$ has the same shape $\vec{\beta}_{2}^{s}$ as $\mat{X}_1^s$,
we should be able to reconstruct $\mat{X}_1^s$ with
\begin{align}
g_\mathrm{dec}\left(\vec{\beta}_{2}^{s}, \vec{\tilde{\theta}}^{t}\right)&=\mat{X}_{1}^{s}.
\label{eq:selfconsistency}
\end{align}
The intuition behind this constraint is that the encoding and decoding of pose code should remain self-consistent with changes in the shape.

Although this loss alone is already quite effective, degeneracy can occur in the network if the proxy mesh $\tilde{\mat{X}}^{t}$ 
inherits shape attributes of $\mat{X}_1^s$ through the pose code. We make sure this does not happen by incorporating ARAP deformation~\cite{sorkine2007rigid} within the training loop.

\subsubsection{As-rigid-as-possible Deformation.}
\begin{figure}[t]
	\centering
	\includegraphics[width=.9\textwidth]{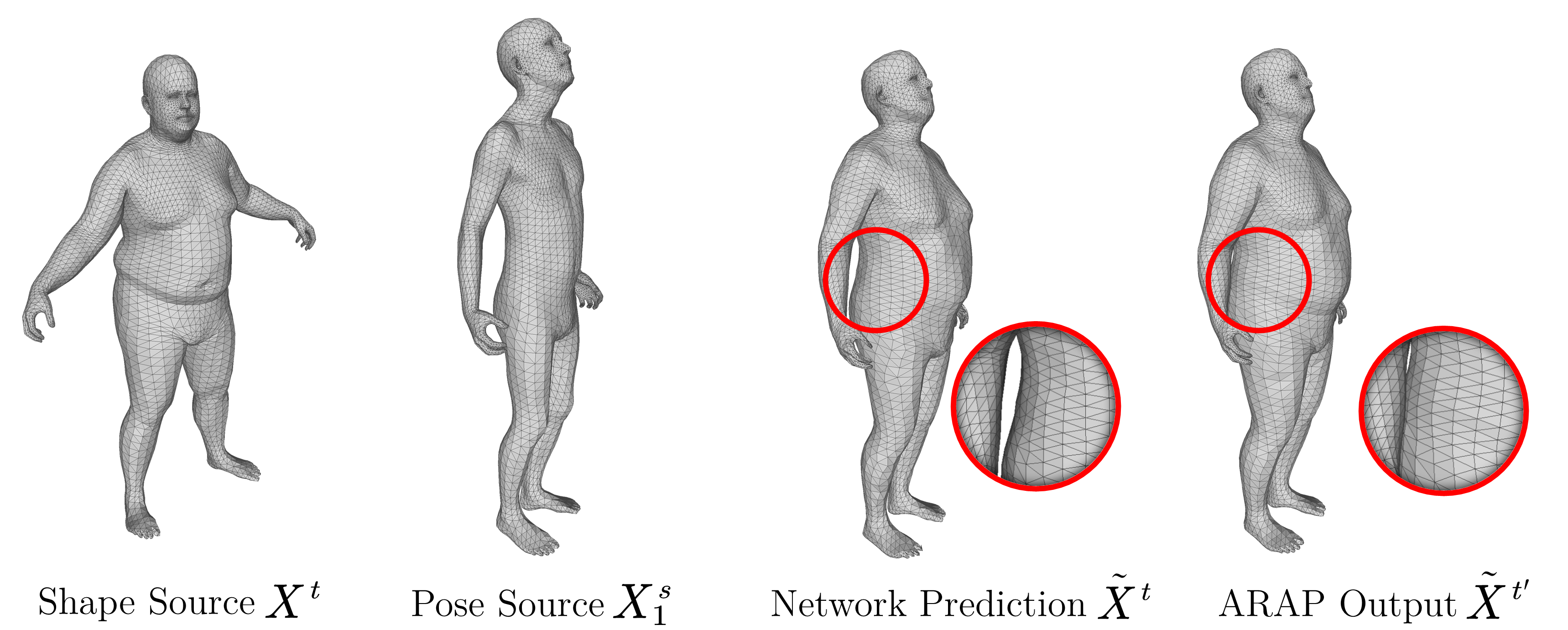}
	\caption{ARAP corrects artifacts in network prediction caused by embedding shape information in pose code. Notice how the circled region in the initial prediction resembles that of the pose source. This is rectified after applying ARAP for only 1 iteration. }
	\label{fig:arap}
\end{figure}

We use ARAP to deform $\mat{X}^{t}$ to match the pose of the network prediction $\mat{\tilde{X}}^{t}$ while preserving the original shape as much as possible,

\begin{equation}
    \mat{\tilde{X}}^{t'} = \text{ARAP}\left(\mat{X}^{t}, \mat{\tilde{X}}^{t}\right),
\end{equation}
where $\mat{\tilde{X}}^{t'}$ is the desired deformed shape, see Fig.~\ref{fig:arap}. Specifically, we deform $\mat{X}^{t}$ to match a few randomly selected anchor points of the network prediction $\mat{\tilde{X}}^{t}$.
ARAP is a detail-preserving surface deformation algorithm that encourages locally rigid transformations. Note that we can successfully apply ARAP because the shape of  $\mat{\tilde{X}}^{t}$ should converge to the shape of $\mat{X}^{t}$ during training. Hence, when only pose is different in the pair $(\mat{X}^{t},\mat{\tilde{X}}^{t})$, the ARAP loss approaches zero, and disentanglement is successful.

In the following, we provide a brief introduction to the optimization procedure of ARAP. We refer interested readers to~\cite{sorkine2007rigid} for more details.
Let $\mat{X}$ be a triangle mesh embedded in $\mathbb{R}^3$ and $\mat{\tilde{X}}$ be the deformed mesh. Each vertex $i$ has an associated cell $\set{C}_i$, which covers the vertex itself and its one-ring neighbourhood $\set{N}(i)$. If a cell $\set{C}_i$ is rigidly transformed to $\set{\tilde{C}}_{i}$, the transformation can be represented by a rotation matrix $\mat{R}_i$ satisfying $\vec{\tilde{e}}_{ij}=\mat{R}_i\vec{e}_{ij}$ for every edge $\vec{e}_{ij} = (\vec{v}_j-\vec{v}_i)$ incident at vertex $\vec{v}_i$.
If $\set{\tilde{C}}_{i}$ and $\set{C}_{i}$ cannot be rigidly aligned, then $\mat{R}_i$ is the optimal rotation matrix that aligns $\set{C}_i$ and $\set{\tilde{C}}_{i}$ with minimal non-rigid distortion.
This objective can be formulated as follows.
% If such a rotation matrix does not exist, $\mat{R}_i$ is then the optimal rotation that approximately aligns $\set{C}_i$ and $\set{C}_i^{'}$. The deviation from rigid transformation is hence measured by
\begin{equation}
E\left( \set{C}_i,\set{\tilde{C}}_{i} \right) =\sum_{j\in \set{N}\left( i \right)}{w_{ij}\left\lVert \vec{\tilde{e}}_{ij}-\mat{R}_i\vec{e}_{ij} \right\rVert ^2}
\label{eq:1}
\end{equation}
where $w_{ij}$ adjusts the importance of each edge.
ARAP deformation minimizes Eq.~\eqref{eq:1} for all vertices $i$ by an iterative procedure. It alternates between first estimating the current optimal rotation $\mat{R}_i$ for cell $\set{C}_i$ while keeping the vertices $\vec{\tilde{v}}_i$ (and hence the edges $\vec{\tilde{e}}_{ij}$) fixed, and second computing the updated vertices $\vec{\tilde{v}}_i$ based on the updated $\mat{R}_i$.
% positions $\vec{\tilde{v}}_i$ of cell $\set{\tilde{C}}_i$.
Let the covariance matrix $\mat{S}_i=\sum_{j\in \set{N}\left( i \right)}{w_{ij}\vec{e}_{ij}\vec{\tilde{e}}_{ij}^{T}}$ have a singular value decomposition,  $\mat{S}_i=\mat{U}_i\mat{\Sigma}
_i\mat{V}_i$. Then the relative rotation $\mat{R}_i$ between them can be analytically calculated as $\mat{R}_i = \mat{V}_i\mat{U}_i^T$ up to a change of sign~\cite{arun1987least}. Fixing $\mat{R}_i$ simplifies Eq.~\eqref{eq:1} to a weighted least squares problem (over the vertices ) of the form
\begin{equation}
    \sum_{j\in \set{N}\left( i \right)}{w_{ij}\left( \vec{\tilde{v}}_i-\vec{\tilde{v}}_j \right)}\,\,=\sum_{j\in \set{N}\left( i \right)}{\frac{w_{ij}}{2}\left( \mat{R}_i+\mat{R}_j \right) \left( \vec{v}_i-\vec{v}_j \right)},
\label{eq:2}
\end{equation}
which can be solved efficiently by a sparse Cholesky solver.

Note that Eq.~\eqref{eq:2} is an underdetermined problem so at least one anchor vertex needs to be fixed to obtain a unique solution. We take $\mat{\tilde{X}}^{t}$ as an initial guess and randomly fix a small number of anchor vertices across its surface that should be matched by deforming the source mesh $\mat{X}^t$ (i.e. $\vec{\tilde{v}}^{t}_j:=\vec{v}^{t}_j$ for all anchor vertices $\vec{v}^{t}_j$).
There is a tradeoff when determining the number of anchor vertices; fixing too many does not improve the shape much while fixing too few could incur a deviation of pose. We found that fixing 1\% - 10\% vertices gives good results in most cases. For training efficiency considerations, we only run ARAP for 1 iteration. This is sufficient since ARAP runs on every input training batch. We also adopted uniform weighting instead of cotangent weighting for $w_{ij}$ and we did not observe any performance drop under this choice.

\subsubsection{Self-consistency Loss.} Let $\mat{\tilde{X}}^{t'}$ be the output of ARAP, which should have the pose of $\mat{X}_1^s$ with the shape of $\mat{X}^t$. We enforce the equality in Eq.~\eqref{eq:selfconsistency} with the following self-consistency loss:
\begin{equation}
\mathcal{L}_S = {\left\lVert {g_\mathrm{dec}\left(f_{\vec{\beta}}(\mat{X}_2^s), f_{\vec{\theta}}(\mathcal{T}(\mat{\tilde{X}}^{t'}))\right) - \mat{X}_1^s} \right\rVert}_{1}
\label{eq:selfconsistencyloss1}
\end{equation}
where again, the intuition is that the pose extracted  $f_{\vec{\theta}}(\mathcal{T}(\mat{\tilde{X}}^{t'}))$ should be independent of shape. 
Note that while ARAP is computed on the fly during training, we do not backpropagate through it.

\subsection{Loss Terms and Objective Function}
The overall objective we seek to optimize is
\begin{equation}
    \mathcal{L} = \lambda_C\mathcal{L}_C + \lambda_S\mathcal{L}_S
\end{equation}
In all our experiments we set $\lambda_C = \lambda_S = 0.5$.
We also experimented with edge length constraints and other local shape preserving losses, but observed no benefit or worse performance. 

\subsection{Implementation Details}
We preprocess the input meshes by centering them around the origin. For the disentangling mesh auto-encoder, we use an architecture similar to~\cite{bouritsas2019neural}. In particular, we adopt the spiral convolution operator, which aggregates and orders local vertices in a spiral trajectory. Each encoder branch consists of four consecutive mesh convolution layers and downsampling layers. The last layer is fully-connected which maps flattened features to latent space. The decoder architecture is a symmetry of the encoder except that mesh downsampling layers are replaced by upsampling layers. We follow the practice in~\cite{ranjan2018generating} which downsamples and upsamples meshes based on quadric error metrics. We choose leaky ReLU with a negative slope of 0.02 as activation function. The model is optimized by ADAM solver with a cosine annealing learning rate scheduler.

\section{Experiments}
In this section, we evaluate our proposed approach on a variety of datasets and tasks. We conduct quantitative evaluations on AMASS dataset and COMA dataset. 
We compare our model to the state-of-the-art unsupervised disentangling models proposed in~\cite{aumentado2019geometric,jiang2019disentangled}. We also perform an ablation study to evaluate the importance of each loss. In addition, we qualitatively show pose transfer results on four datasets (AMASS, SMAL, COMA and MANO) to demonstrate the wide applicability of our method. Finally, we show the usefulness of our disentangled codes for the tasks of shape and pose retrieval and motion sequence interpolation.

\subsection{Datasets}
We use the following four publicly available datasets to evaluate our method:
\textbf{AMASS}~\cite{mahmood2019amass} is a large human motion sequence dataset that unifies 15 smaller datasets by fitting SMPL body model to motion capture markers. It consists of 344 subjects and more than 10k motions. We follow the protocol splits and sample every 1 out of 100 frames for the middle 90\% portion of each sequence.

\noindent \textbf{SMAL}~\cite{zuffi20173d} is a parametric articulated body model for quadrupedal animals. Since there are not sufficient scans in this dataset, we synthesize SMAL shapes and poses using the procedure in~\cite{groueix20183d}. Finally, we get 100 shapes and 160 poses distinct for each shape. We use a 9:1 data split.

\noindent \textbf{MANO}~\cite{romero2017embodied} is the 3D hand model used to fit AMASS together with SMPL. We treat it as a standalone dataset since its training scans contain more pose variations. To keep things simple without losing generality, we train the model specifically on right hands and flipped left hands. The official training set contains less than 2000 samples, hence we augment it by sampling shape and pose parameters of MANO from a Gaussian distribution.

\noindent \textbf{COMA}~\cite{ranjan2018generating} is a facial expression dataset consisting of 12 subjects under 12 types of extreme expressions. We follow the same splits as in~\cite{ranjan2018generating}.

\subsection{Quantitative Evaluation}
\subsubsection{AMASS Pose Transfer}
In the following, we show quantitative results of our model trained on AMASS. Since AMASS comes with SMPL parameters, we utilize the SMPL model to generate pseudo-groundtruth for evaluating pose-transferred reconstructions. We sample a subset of paired meshes (with different shapes and poses) along with their pose-transferred pseudo-groundtruth. The error is calculated between model-predicted transfer results and the pseudo-groundtruth. We use $128$-dimensional latent codes, $16$ for shape and $112$ for pose. 

We compare our method to Geometric Disentanglement Variational Autoencoder(GDVAE)~\cite{aumentado2019geometric}, a state-of-the-art unsupervised method which can disentangle pose and shape from 3D pointclouds. It is important to note that a fair comparison to GDVAE is not possible as we make different assumptions. They do not assume mesh correspondence while we do. However, GDVAE uses LBO spectra computed on meshes which are in perfect correspondence. Since the LBO spectra is sensitive to noise and the type of discretization, the performance of GDVAE could be significantly deteriorated when computed on meshes not in correspondence. Furthermore, we assume we can see the same shape in different poses. But as argued earlier, this is the typical case in datasets with dynamics. 
Hence, despite the differences in assumptions, we think the comparison is meaningful.

We report the one-side Chamfer distance for GDVAE (\emph{i.e.}, average distance between every point and its nearest point on groundtruth surface) and report the vertex-to-vertex error for our method. Note that the Chamfer distance would be lower for our method, but we want the metric to reflect how well we predict the semantics (body part locations) as well. 

We also compare our method with a supervised baseline, which leverages pose labels from the SMPL. In that case, the intermediate mesh $\mat{\tilde{X}}^{t'}$ is replaced by the pseudo-groundtruth coming from the SMPL model.
\begin{table}[t]
\centering
\begin{tabular}{c|c|c|c|c|c}
 &  GDVAE&  \makecell{Ours \\ with ARAP}& \makecell{Ours \\ without ARAP} & \makecell{Ours \\ without self-consistency} &\makecell{Ours \\ supervised}\\
\hline
Mean Error &  54.44& 19.43 & 20.27 & 23.83& 15.44
\end{tabular}
\caption{AMASS pose transfer results when training on different models. The numbers are measured in millimeters. The error on our model is close to the supervised baseline, indicating that our self-consistency loss is a good substitute for pose supervision.}
\label{tab:amass}
\end{table}

Table~\ref{tab:amass} summarizes reconstruction errors of pose-transferred meshes on AMASS dataset using different models. The supervised baseline with pose supervision achieves the lowest error, which serves as the performance upper bound for our model. Remarkably, our unsupervised model is only $4 mm$ worse than the supervised baseline, suggesting that our proposed approach, which only requires seeing a subject in different poses, is sufficient to disentangle shape from pose. 
In addition, our approach achieves a much lower error compared to GDVAE. Again, we compare for completeness, but we do not want to claim we are superior as our assumptions are different, and the losses are conceptually very different. 

We can also observe from Table~\ref{tab:amass} that training solely with cross-consistency constraint leads to degenerate solutions. 
This shows that our approach can only exploit the weak signal of seeing the same subject in different poses when combined with the self-consistency loss.
Notably, enforcing the self-consistency constraint already drives the model to learn a reasonably well-disentangled representation, which is further improved by incorporating ARAP in-the-loop. We hypothesize that without ARAP, the intermediate mesh $\tilde{X}^t$ is noisy in shape but relatively accurate in pose at early stages of training, thus helping disentanglement. 

\subsubsection{AMASS Pose-aware Shape Retrieval}
Shape retrieval refers to the task of retrieving similar objects given a query object. Our model learns disentangled representations for shape and pose; hence we can retrieve objects either similar in shape or similar in pose. Our evaluation of shape retrieval accuracy follows the experiment settings in~\cite{aumentado2019geometric}. Specifically, we evaluate on AMASS dataset which comprises groundtruth SMPL parameters. To avoid confusion of notations, we denote with $\vec{\dot\beta}$ the SMPL shape parameters and denote with $\vec{\dot\theta}$ the SMPL pose parameters. For each queried object $\mat{X}$, we encode it into a latent code and search for its closest neighbour $\mat{Y}$ in latent space. The retrieval accuracy is determined by the Euclidean error between SMPL parameters of $\mat{X}$ and $\mat{Y}$: $E_{\vec{\dot\beta}}(\mat{X}, \mat{Y}) = \lVert \vec{\dot\beta}(\mat{X}) - \vec{\dot\beta}(\mat{Y}) \rVert_2$, $E_{\vec{\dot\theta}}(\mat{X}, \mat{\mat{Y}}) = \lVert q(\vec{\dot\theta}(\mat{X})) - q(\vec{\dot\theta}(\mat{Y})) \rVert_2$, where $q(\cdot)$ converts axis-angle representations to unit quaternions. Again, to properly compare with GDVAE which uses 5 dimensions for shape and 15 dimensions for pose, we reduce the latent dimension of our model with principal component analysis(PCA). We show results for shape retrieval and pose retrieval in Table~\ref{tab:retrieval}.

\begin{figure}[t]
\centering
\includegraphics[width=.8\textwidth]{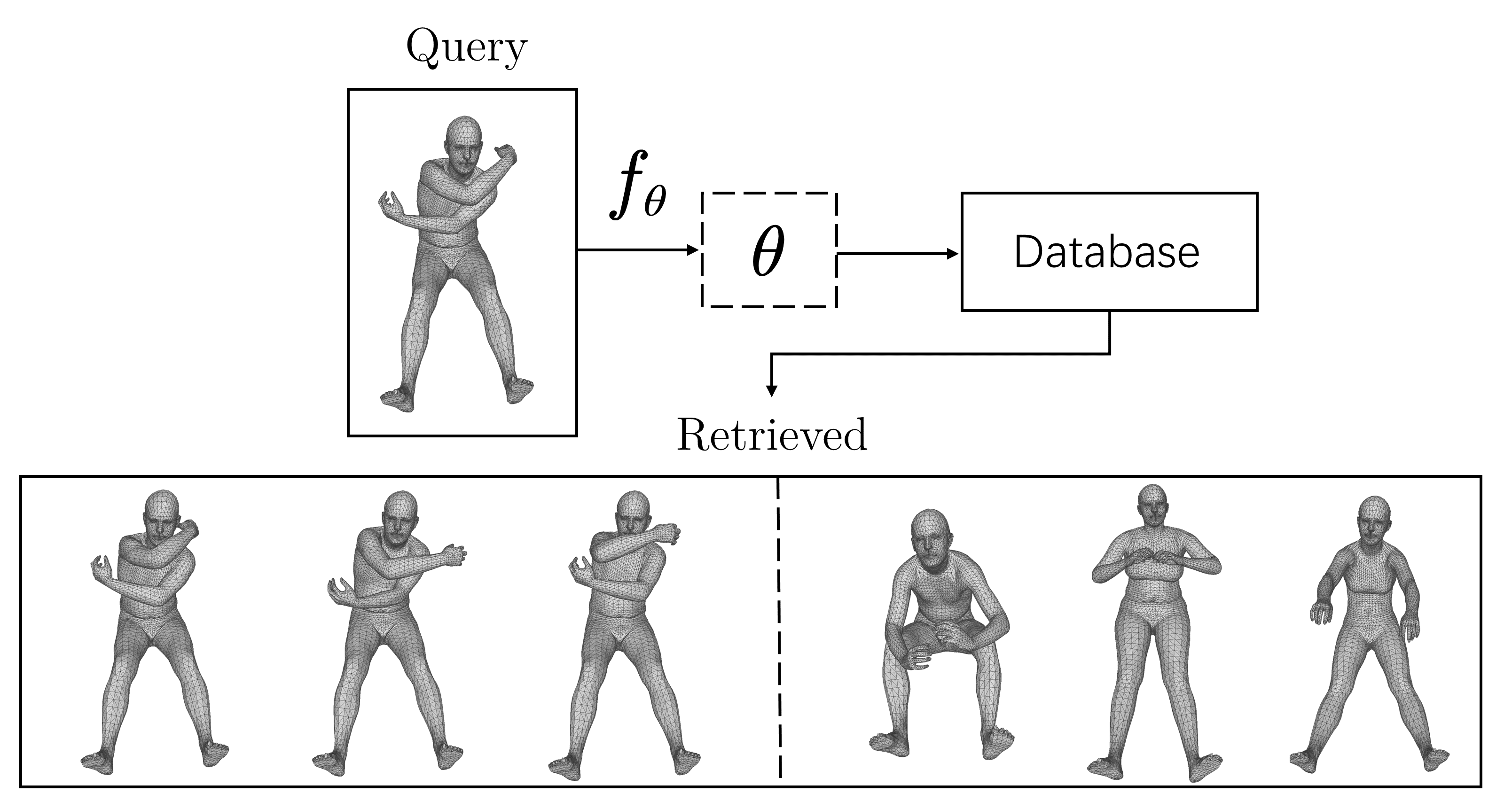}
\caption{An example of pose retrieval with our model. Bottom left: top three meshes most similar with the query in pose code. Bottom right: top three meshes of \emph{different subjects} most similar with the query in pose code.}
\label{fig:retrieval}
\end{figure}

Ideally if the shape code is disentangled from the pose code, we should get a low $E_{\vec{\dot\beta}}$ and high $E_{\vec{\dot\theta}}$ when retrieving with $\vec{\beta}$, and vice versa. This is in accordance with our results. Interestingly, dimensionality reduction with PCA boosts the shape difference for pose retrieval. This indicates that some degree of entanglement is still present in our pose code. An example of pose retrieval is demonstrated in Fig.~\ref{fig:retrieval} -- notice the pose similarity for the retrieved shapes.
\begin{table}[t]
\centering
\begin{tabular}{>{\centering}p{0.22\textwidth}>{\centering} p{0.1\textwidth}|>{\centering}p{0.08\textwidth}|p{0.08\textwidth}<{\centering}}
\hline
& & $\vec{\beta}$ & $\vec{\theta}$ \\
\hline
\multirow{2}{*}{GDVAE} & $E_{\vec{\dot\beta}}$ &2.80 $\downarrow$ & 4.71 $\uparrow$ \\
                       & $E_{\vec{\dot\theta}}$ &1.47 $\uparrow$ & 1.44  $\downarrow$ \\
\hline
\multirow{2}{*}{Ours - with PCA} & $E_{\vec{\dot\beta}}$ &0.34 $\downarrow$ & 2.14 $\uparrow$ \\
                      & $E_{\vec{\dot\theta}}$ &1.23 $\uparrow$ & 0.87 $\downarrow$\\
\hline
\multirow{2}{*}{Ours - without PCA} & $E_{\vec{\dot\beta}}$ &0.14 $\downarrow$ & 0.92 $\uparrow $ \\
                      & $E_{\vec{\dot\theta}}$ &0.94 $\uparrow$ & 0.76 $\downarrow$\\
\hline
\\
\end{tabular}
\caption{Mean error on SMPL parameters for shape retrieval. Column 1 corresponds to retrieval with shape code $\vec{\beta}$ and column 2 with pose code $\vec{\theta}$.
	Arrows indicate if the desired metrics should be high or low when retrieving with $\vec{\beta}$ or $\vec{\theta}$.
	 \vspace{-0.5cm}}
\label{tab:retrieval}
\end{table}
\subsubsection{COMA Expression Extrapolation}
COMA dataset spans over twelve types of extreme expressions. To evaluate the generalization capability of our model, we adopt the expression extrapolation setting of~\cite{ranjan2018generating}. Specifically, we run a 12-fold cross-validation by leaving one expression class out and training on the rest. We subsequently evaluate reconstruction on the left-out class.
\begin{table}[t]
\centering
\begin{tabular}{ >{\centering}p{0.13\textwidth} | >{\centering}p{0.13\textwidth} |>{\centering}p{0.16\textwidth} | p{0.13\textwidth}<{\centering}}
 &  Ours&  Jiang et al.'s&FLAME\\
\hline
average & \textbf{1.28}& 1.64&  2.00 \\
\end{tabular}
\caption{Mean errors of expression extrapolation on COMA dataset. All numbers are in millimeters. The results of Jiang et al. and FLAME are taken from~\cite{jiang2019disentangled}.}
\label{tab:coma}
\end{table}
Table~\ref{tab:coma} shows the average reconstruction performance of our model compared with FLAME~\cite{FLAME:2017} and Jiang et al.'s approach~\cite{jiang2019disentangled} (see supplementary material for the full table). Both Jiang et al. and our model allocate 4 dimensions for identity and 4 dimensions for expression, while FLAME allocates 8 dimensions for each. Our model consistently outperforms the other two by a large margin.

\subsection{Qualitative Evaluation}
\subsubsection{Pose Transfer}
\begin{figure}[t]
\centering
\includegraphics[width=\textwidth]{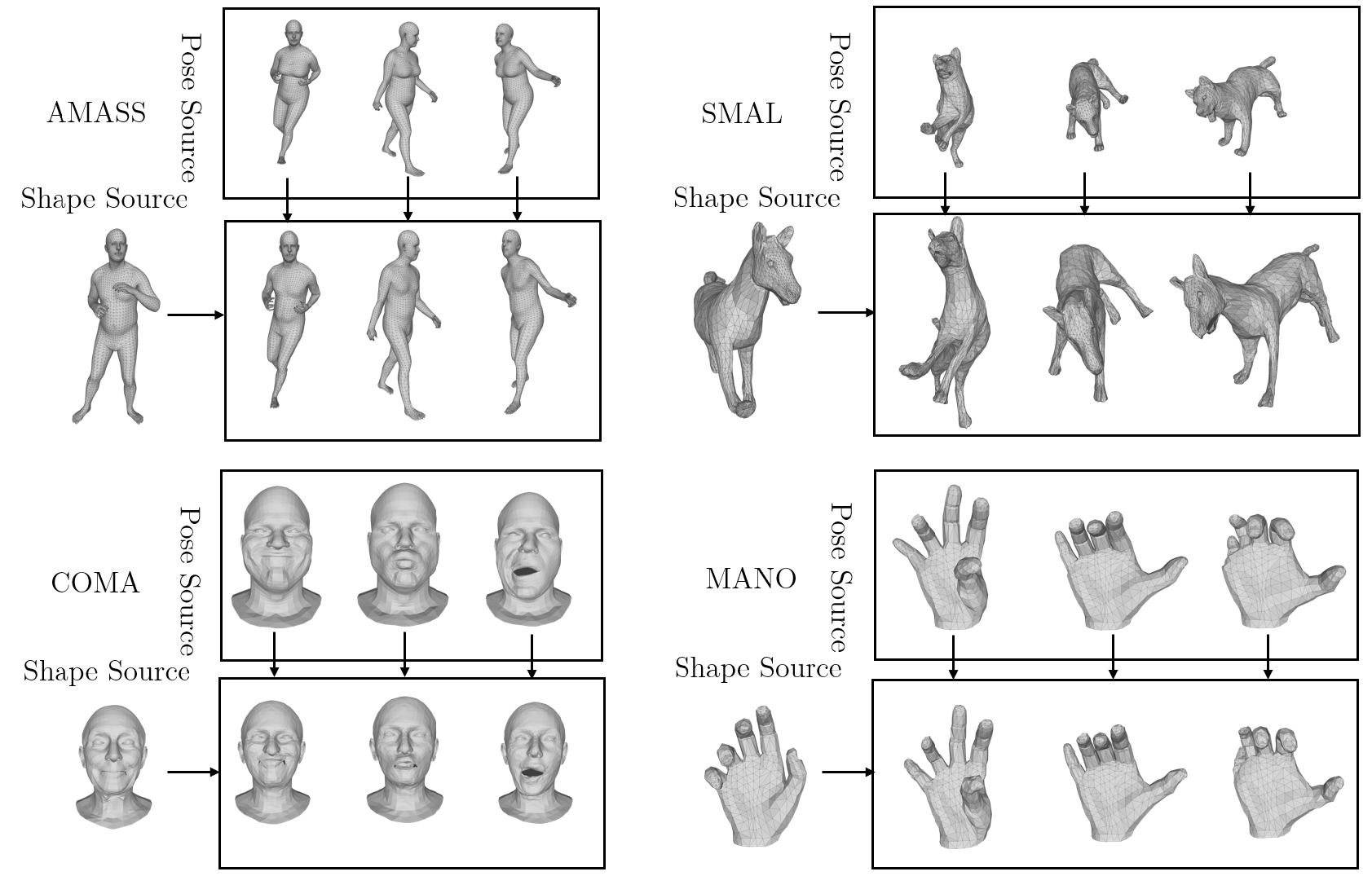}
\caption{Pose transfer from pose sources to shape sources. Please see supplementary video at \url{https://virtualhumans.mpi-inf.mpg.de/unsup_shape_pose/} for transferring animated sequences.}
\label{fig:transfer}
\end{figure}

We qualitatively evaluate pose transfer on AMASS, SMAL, COMA and MANO. In each dataset, a pose sequence is transferred to a given shape. Ideally if our model learns a disentangled representation, the outputs should preserve the identity of shape source, while inheriting the deformation from pose sources. Fig.~\ref{fig:transfer} visualizes the transfer results. We can observe subject shape is preserved well under new poses. The results are most obvious for bodies, animals and faces. It is less obvious for hands due to their visual similarity.
\begin{figure}[t]
	\centering
	\includegraphics[width=\textwidth]{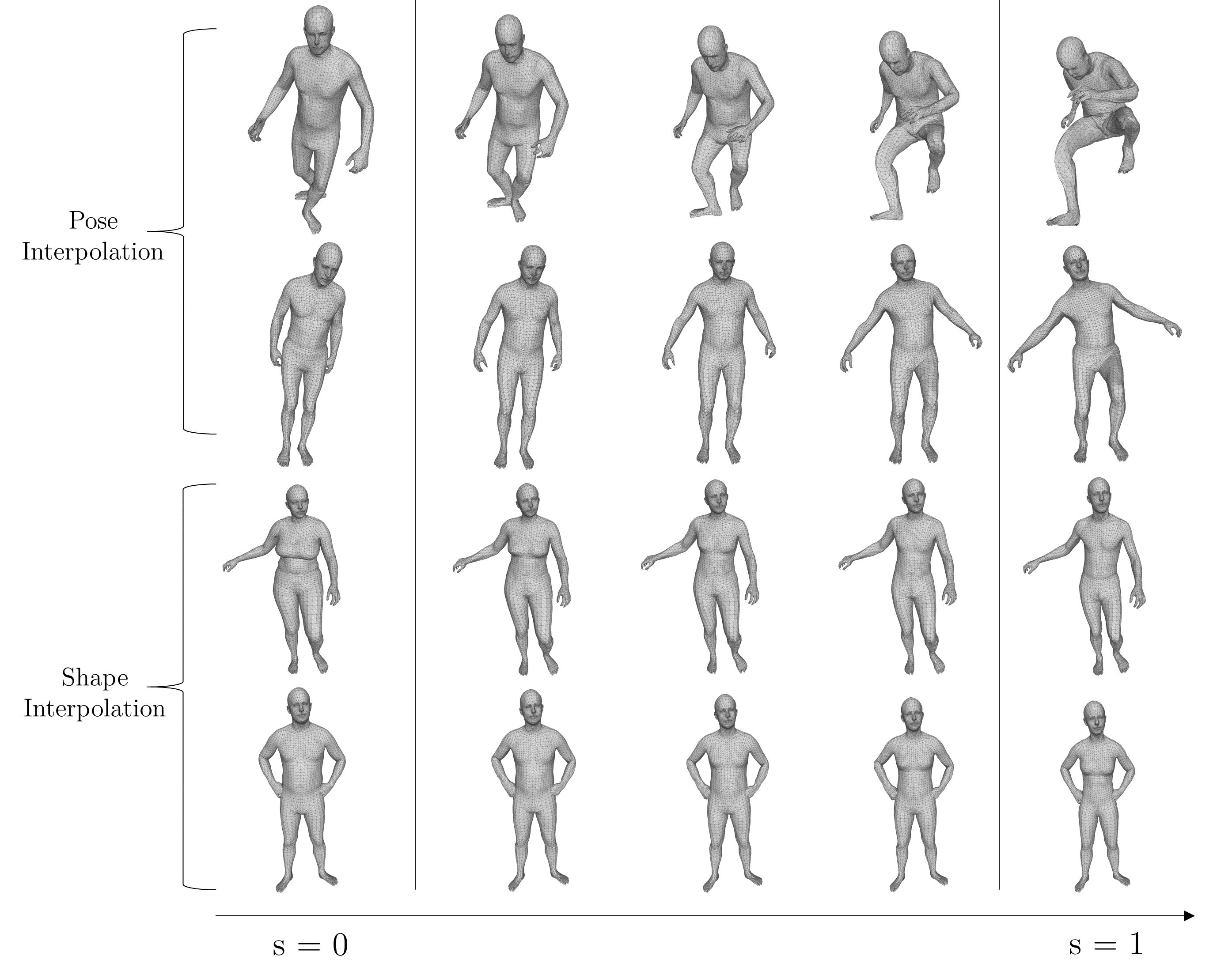}
	\caption{Latent interpolation of shape and pose codes on AMASS dataset. The leftmost column are source meshes, while the rightmost are target meshes. Intermediate columns are linear interpolation of specific codes at uniform time steps $s=0$ and $s=1$. First
two rows show interpolation of pose, and last two rows show interpolation of shape.}
	\label{fig:interp}
\end{figure}
\subsubsection{Latent Interpolation}
Latent representations learned by our model should ideally be smooth and vary continuously. We demonstrate this via linearly interpolating our learned shape codes and pose codes. When interpolating shape, we always fix the pose code to that of the source mesh. The same holds when we interpolate pose. Interpolation results are shown in Fig.~\ref{fig:interp}. We can observe the smooth transition between nearby meshes. Furthermore, we can see that mesh shapes remain unchanged during pose interpolation, and vice versa. This indicates that variations in shape and pose are independent of each other.
\section{Conclusion and Future Work}
In this paper, we introduced an auto-encoder model that disentangles shape and pose for 3D meshes in an unsupervised manner. We exploited subject identity information, which is commonly available when scanning or capturing shapes using motion capture. 
We showed two key ideas to achieve disentanglement, namely a cross-consistency and a self-consistency loss coupled with ARAP deformation within the training loop. 
Our model is straightforward to train and it generalizes well across various datasets. We demonstrated the use of latent codes by performing pose transfer, shape retrieval and latent interpolation. 
Although our method provides an exciting next step in unsupervised learning of deformable models from data, there is still room for improvement. In contrast to hand-crafted models like SMPL, where every parameter carries meaning (joint axes and angles per part), we have no control over specific parts of the mesh with our pose code. 
We also observed that interpolation of large torso rotations squeezes the meshes. In future work, we plan to explore a more structured pose space for easier part manipulation, which allows easy user manipulation, and plan to generalize our method to work with un-registered pointclouds as input.
Since our model builds on simple yet effective ideas, we hope researchers can build on it and make further progress in this exciting research direction. 

\subsubsection{Acknowledgements.}This work is funded by the Deutsche Forschungsgemeinschaft (DFG,
German Research Foundation) - 409792180 (Emmy Noether Programme,
project: Real Virtual Humans). We also want to thank members of Real Virtual Humans group for useful discussion.

\bibliographystyle{splncs04}
\bibliography{egbib}
\end{document}